\documentclass[letterpaper, 10 pt, conference]{ieeeconf}  

\IEEEoverridecommandlockouts                              
\usepackage{amsmath}
\usepackage{graphicx}
\usepackage{url}
\usepackage{hyperref}

\begin{document}




\title{\LARGE \bf
Web-Gewu: A Browser-Based Interactive Playground for Robot Reinforcement Learning
}

\author{Kaixuan Chen, Linqi Ye
\thanks{*This work was supported by the Science and Technology Commission of Shanghai Municipality (Grant No.~24511103304).}%
\thanks{The authors are with the School of Future Technology, Shanghai University, Shanghai 200444, China. 
        (Corresponding author: Linqi Ye, email: yelinqi@shu.edu.cn)}%
}


\maketitle
\thispagestyle{empty}
\pagestyle{empty}

\begin{abstract}

With the rapid development of embodied intelligence, robotics education faces a dual challenge: high computational barriers and cumbersome environment configuration. Existing centralized cloud simulation solutions incur substantial GPU and bandwidth costs that preclude large-scale deployment, while pure local computing is severely constrained by learners' hardware limitations. To address these issues, we propose \href{http://47.76.242.88:8080/receiver/index.html}{Web-Gewu}, an interactive robotics education platform built on a WebRTC cloud-edge-client collaborative architecture. The system offloads all physics simulation and reinforcement learning (RL) training to the edge node, while the cloud server acts exclusively as a lightweight signaling  relay, enabling extremely low-cost browser-based peer-to-peer (P2P) real-time streaming. Learners can interact with multi-form robots at low end-to-end latency directly in a web browser without any local installation, and simultaneously observe real-time visualization of multi-dimensional monitoring data, including reinforcement learning reward curves. Combined with a predefined robust command communication protocol, Web-Gewu provides a highly scalable, out-of-the-box, and barrier-free teaching infrastructure for embodied intelligence, significantly lowering the barrier to entry for cutting-edge robotics technology.

\end{abstract}

\section{INTRODUCTION}

With the explosive growth of embodied intelligence and robotics, complex dynamic robot control is entering the public consciousness at an unprecedented pace. However, the cutting-edge algorithms underpinning these systems---such as reinforcement learning (RL) and model predictive control---remain opaque to most students and educators. This cognitive gap creates strong demand for intuitive, accessible tools that can help non-specialist learners understand the core logic of these algorithms.

Integrating robotics into education is an effective approach, but the existing technological ecosystem erects significant barriers. On one hand, mainstream RL simulation platforms such as Isaac Lab~\cite{c1}, mjlab~\cite{c2}, Genesis, and Gewu Playground~\cite{c3} impose stringent local GPU requirements or complex environment configuration, directly excluding learners with limited hardware resources and exacerbating inequality in access to cutting-edge technology. On the other hand, traditional centralized cloud simulation solutions, while bypassing local hardware limitations, incur prohibitive GPU and bandwidth costs, making large-scale deployment in universal teaching settings economically infeasible.

\begin{figure}[thpb]
  \centering
  \includegraphics[width=\columnwidth]{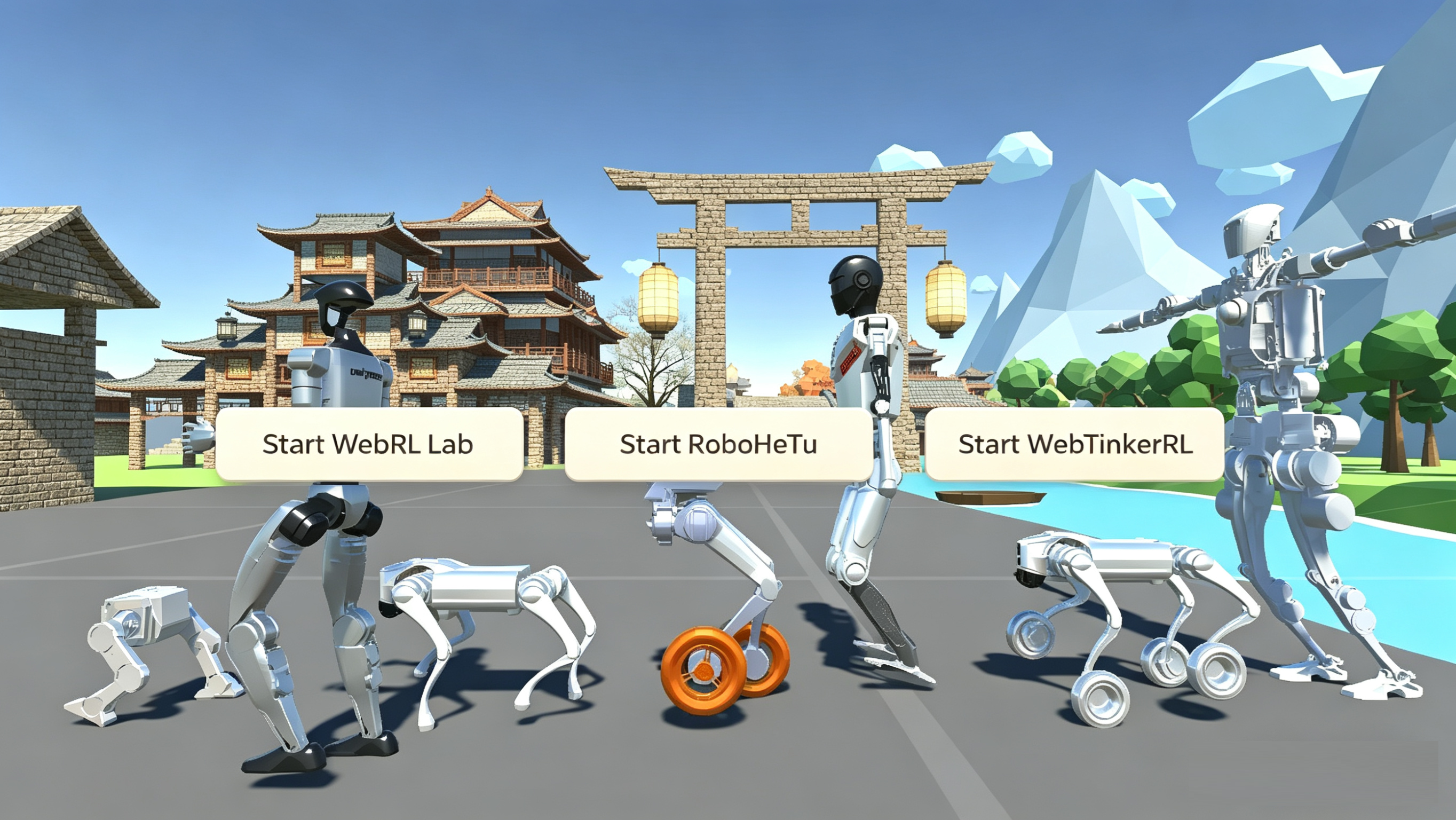}
  \caption{Main interface of Web-Gewu. It can be visited via \url{http://47.76.242.88:8080/receiver/index.html}.}
  \label{fig:ui}
\end{figure}

To break down this computational barrier and architectural dilemma, we present Web-Gewu (Figure \ref{fig:ui}): an interactive robotics education sandbox built on a cloud-edge-client collaborative WebRTC~\cite{webrtc} architecture. We innovatively restructure the computational topology by offloading all resource-intensive physics simulation and RL training entirely to the edge node, while the cloud server acts solely as a lightweight signaling relay to facilitate NAT traversal. Learners enjoy a truly zero-installation experience---no local GPU configuration or physical robot hardware is required---and simply open a browser to stream the simulation in real time via low-cost P2P connectivity.

Web-Gewu provides a purely virtual interactive environment with multiple built-in RL demonstration tasks. Through a low-latency streaming mechanism, the system transmits training and inference processes from the edge node to the client in real time, transforming the RL training process into intuitive visual feedback on the screen. Users can intervene in the behavior of bipedal, quadrupedal, and leg-wheeled robots in real time through a standard web browser with no perceptible lag.

The core contributions of this paper are as follows:
\begin{enumerate}
\item We develop Web-Gewu, a cross-platform, purely browser-based robot education platform that supports real-time control and gait verification of multi-form robots (bipedal, quadrupedal, and leg-wheeled) without any installation.
\item We propose a lightweight cloud-edge-client simulation architecture in which the edge node handles all heavy computation and the cloud provides only signaling relay, breaking the cost bottleneck of traditional high-fidelity cloud rendering and enabling universal low-latency remote interaction.
\item We construct a dynamic visualization closed loop for RL algorithms by transmitting multi-dimensional monitoring data (reward curves, curriculum learning stages) in real time through the WebRTC DataChannel, providing learners with intuitive insight into the data-driven nature of reinforcement learning.
\item We establish a minimal scene integration contract within the layered framework: a new robot scenario can be onboarded without touching any WebRTC, signaling, or protocol code---it requires only a Unity scene runtime module, a \texttt{SceneDirector} registration, and at most one new typed command handler, making the platform straightforwardly extensible by the research community.
\end{enumerate}

\section{RELATED WORK}

\subsection{Interactive Educational Tools for AI Learning}

The web-based architecture inherently offers zero-deployment, cross-platform, and instant-interaction advantages, making it an excellent medium for lowering AI education barriers. Lightweight interactive tools such as TensorFlow Playground~\cite{c4}, Reinforcement Learning Playground~\cite{rl_playground}, and Interactive Deep RL Demo \cite{interactive_deeprl_demo} have successfully popularized deep learning and RL concepts through intuitive 2D visualization in the browser, demonstrating the enormous potential of the web as a universal medium for AI education.

However, such systems are typically designed for low-dimensional or abstract tasks, focusing on algorithmic explanation rather than modeling real-world physical environments. The core of robot RL lies in the complex interaction between an agent and a continuous physical world, requiring a high-fidelity 3D physics engine for accurate rigid-body dynamics, collision feedback, friction, and joint constraint calculations. Traditional web tools, constrained by browser compute limitations, can only handle extremely simple tasks and cannot support the large-scale parallel sampling or high-frequency control loops required for robot RL. This creates a persistent structural contradiction between the extreme accessibility of the web and the high computational demands of robotic systems---a gap that Web-Gewu is designed to bridge.

\subsection{Robot Simulation and RL Platforms}

Debugging robot systems in a real physical environment is not only time-consuming but also incurs high trial-and-error costs and hardware damage risks. Building a high-fidelity, high-throughput simulation environment is therefore a prerequisite for converging robot RL algorithms.

Early robot simulation platforms such as Webots~\cite{c5}, Gazebo~\cite{c6}, and PyBullet advanced robot control and kinematics research significantly over the past decade. However, these platforms rely on CPU-based serial or limited-parallel physics solvers, revealing significant computational bottlenecks when large-scale environmental interaction data is required for RL training.

To overcome this, modern platforms such as Isaac Lab, mjlab, and Genesis employ GPU-accelerated physics tensor parallelism or highly optimized C++ back-ends to achieve massive-scale concurrent training of thousands of agents, compressing training time from weeks to hours. However, their closed rendering pipelines, stringent low-level system dependencies, and limited cross-platform compatibility make them unsuitable as universally accessible interactive tools for general education.

Gewu Playground~\cite{c3}, as a simulation foundation deeply integrated with the Unity ML-Agents Toolkit~\cite{mlagents1}~\cite{mlagents2}, supports standard RL routines including complex whole-body imitation and omnidirectional locomotion. Researchers have successfully reproduced the Adversarial Motion Priors (AMP)~\cite{c7} algorithm within Unity, demonstrating its readiness for high-difficulty robot RL. Critically for our work, Unity's built-in WebRTC/Render Streaming pipeline enables high-fidelity simulation rendering to be streamed to a web browser with minimal overhead, while its rich ecosystem---covering sensor simulation, complex terrain generation, and URDF import for robotic systems---significantly reduces the engineering costs of environment setup. This integration of high-fidelity physics simulation with commercial-grade rendering and web streaming is the core enabler of Web-Gewu's out-of-the-box, low-latency browser experience.

\section{SYSTEM ARCHITECTURE}

The Web-Gewu framework implements a low-latency cloud rendering and remote operation paradigm specifically designed for embodied intelligence education. By deeply decoupling heavy physics simulation from the client interface, the system maintains high-fidelity visual feedback while ensuring reliable bidirectional control communication across diverse network topologies.

\subsection{Cloud-Edge-Client Topology}

\begin{figure}[thpb]
  \centering
  \includegraphics[width=\columnwidth]{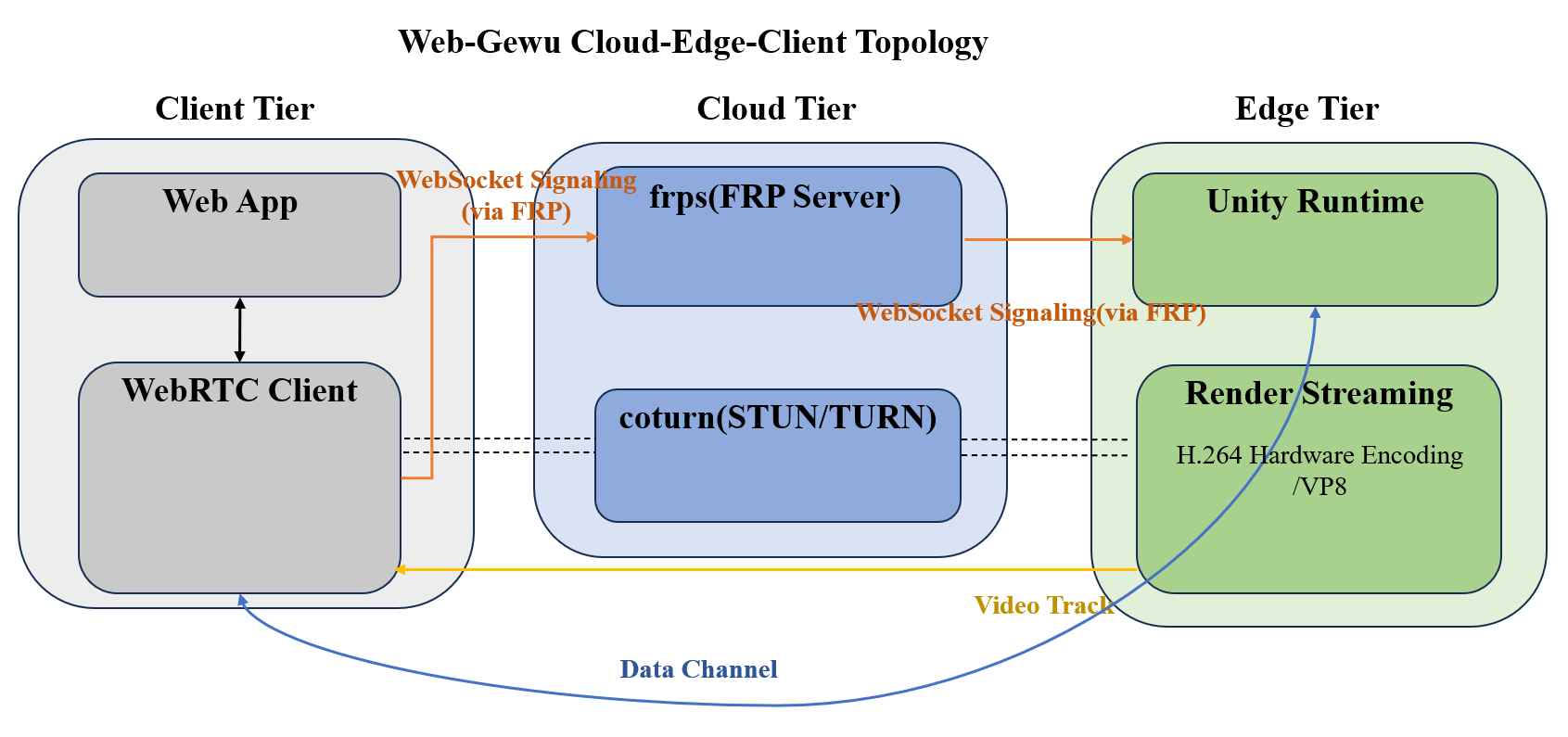}
  \caption{The Cloud-Edge-Client topology of the Web-Gewu framework. The Edge Tier acts as the computation singularity, encapsulating high-fidelity physics simulation and H.264 hardware-accelerated render streaming. The Cloud Tier provides lightweight signaling relay (via FRP) and NAT traversal (via STUN/TURN). The Client Tier functions as a thin web interface, establishing low-latency P2P media (VideoTrack) and bidirectional control (DataChannel) pipelines.}
  \label{fig:architecture}
\end{figure}

To optimize computational resource allocation and minimize end-to-end latency, Web-Gewu adopts a three-tier decoupled topology, as illustrated in Fig.~\ref{fig:architecture}.

\textbf{Edge Tier (Compute Layer).} The edge node---a local GPU-equipped workstation---hosts all computationally intensive tasks: the Unity runtime, high-fidelity rigid-body physics simulation, RL training and inference logic, and hardware-accelerated WebRTC media encoding via Unity Render Streaming. No rendering or business logic is processed in the cloud.

\textbf{Cloud Tier (Signaling Layer).} A lightweight public cloud server is strictly restricted to signaling relay and NAT traversal support. It runs \texttt{frps} for reverse-proxy port mapping (exposing the edge-side Node.js signaling server on a public address) and \texttt{coturn} for STUN/TURN services. A standard 2-core vCPU instance with 2~GB RAM suffices for this role, sustaining CPU utilization of only 0.8\%--1.1\% with a single active client, demonstrating the economic viability of large-scale deployment.

\textbf{Client Tier (Browser Layer).} The learner accesses the system through a standard web browser with no installation required. Even a mobile device can seamlessly dispatch commands to the full edge computing stack. The browser-based WebApp handles UI interaction, command dispatch, and real-time telemetry visualization.

\subsection{Low-Latency Render Streaming Pipeline}

Web-Gewu leverages WebRTC~\cite{c8} as the unified transport layer for both video streaming and bidirectional control. Unity serves as the sole rendering authority: the Render Streaming plugin captures the designated simulation camera, encodes each frame via hardware-accelerated H.264 (with VP8 as a fallback), and publishes it as a WebRTC \texttt{VideoTrack}. The browser-side client binds the decoded frames directly to an HTML5 \texttt{<video>} element for immediate display; no scene reconstruction occurs on the client side.

Connection establishment follows the standard ICE (Interactive Connectivity Establishment) procedure. Both the edge node and the browser independently gather ICE candidates and query the \texttt{coturn} STUN server to resolve their reflexive public-network addresses. The ICE agent then attempts a direct P2P UDP connection through NAT hole-punching. Under favorable network conditions, all media and control traffic flows directly between the edge and the client, producing end-to-end visual latency imperceptible to the operator and eliminating the motion-sickness-inducing lag of conventional remote-desktop approaches. When symmetric NAT or firewall policies prevent direct connectivity, the TURN relay in \texttt{coturn} transparently handles the session, maintaining 100\% connection success across heterogeneous network environments while keeping the media path entirely independent of the cloud signaling infrastructure.

\subsection{Robust Bidirectional Control via Envelope Protocol}

The same WebRTC session that carries the simulation video also hosts a bidirectional \texttt{RTCDataChannel} named \texttt{input}. All control commands originating from the browser and all telemetry responses from Unity traverse this single logical channel, enabling tight synchronization between user actions and system state feedback.

To impose semantic structure and robustness on the message stream, Web-Gewu defines the \emph{Envelope JSON} protocol. Each message is wrapped in a typed envelope carrying a protocol version (\texttt{v}), a globally unique identifier (\texttt{id}), a semantic type string (\texttt{type}), a sender identifier (\texttt{source}), and a precise timestamp (\texttt{ts}), as shown below:

\begin{verbatim}
{ "v":1, "id":"env-m9x2...",
  "type":"scene.load", "source":"web",
  "ts":1710000000000,
  "payload":{ "scene":"RoboHeTu" } }
\end{verbatim}

The \texttt{type} field drives dispatching on both ends. Web-Gewu distinguishes two semantic command categories to maximize robustness under unreliable networks.

\textit{State-intent commands} (e.g., \texttt{scene.load}, \texttt{training.set\_flag}) express a desired system state rather than a strictly ordered action sequence. The Unity-side \texttt{SceneDirector} deduplicates these by ignoring requests for a scene already in the process of loading, preventing cascade failures from network jitter or repeated clicks.

\textit{Continuous control snapshots} (e.g., robot movement direction and speed) represent the current instantaneous control state. Since each snapshot inherently supersedes all prior ones, the receiver discards stale packets and applies only the most recent state, naturally tolerating packet loss without retransmission overhead and eliminating control stream collapse under network backpressure.

The browser-side \texttt{RtcChannelClient} maintains an outbound buffer that retains commands issued before the DataChannel is open, flushing them atomically upon channel establishment. On the Unity side, all inbound WebRTC callbacks are enqueued onto a Unity main-thread processing queue and consumed sequentially in each \texttt{Update()} frame, preventing cross-thread scene mutations and ensuring command ordering.

\subsection{Layered Engineering Architecture}

Web-Gewu's implementation is organized into five decoupled layers to achieve fault isolation, component reusability, and low-cost scenario extensibility:

\begin{enumerate}
\item \textbf{Transport Layer} (\texttt{rtc-channel.js} / \texttt{WebRtcModelCommandBridge.cs}): manages DataChannel lifecycle, JSON serialization, pre-connection buffering, and Unity main-thread scheduling. Scene logic is fully insulated from ICE/SDP details.
\item \textbf{Protocol Layer} (\texttt{envelope.js} / \texttt{Runtime/Handlers/}): defines message formats and dispatches incoming envelopes by \texttt{type}, maintaining backward compatibility with legacy command formats.
\item \textbf{Orchestration Layer} (\texttt{SceneDirector.cs}): manages scene loading, alias resolution, active-scene transitions, camera synchronization, deferred command replay, and in-flight deduplication.
\item \textbf{Scene Runtime Layer} (e.g., \texttt{G1moeAgent}, \texttt{TinkercoinAgent}): executes physics simulation, RL training and inference, and produces typed telemetry payloads.
\item \textbf{UI Interaction Layer} (\texttt{scene-controller.js}): maps user gestures to semantic envelope payloads and renders incoming telemetry; it carries no Unity business logic.
\end{enumerate}

This layered design deliberately minimizes the integration surface for new simulation scenarios. A new scene author need not understand ICE candidate gathering, SDP negotiation, DataChannel open timing, or browser compatibility quirks---those concerns are fully absorbed by the Transport and Protocol layers. In concrete terms, onboarding a new scenario requires satisfying only four contracts: (1) register the scene with \texttt{SceneDirector} so it can be loaded and aliased; (2) expose a bindable runtime object through which the Orchestration Layer can forward commands; (3) reuse existing Envelope \texttt{type} strings where semantics match, or add a single new typed handler if they do not; and (4) optionally define a telemetry payload for any data the scene wishes to stream back to the browser. Crucially, the typed-dispatch design means extending the protocol never requires rewriting a monolithic command parser---it is strictly additive. Scene lifecycle concerns such as deduplication of repeated load requests, deferred command replay while a scene is still initializing, and camera synchronization on transition are all handled by \texttt{SceneDirector} unconditionally, so new scenes inherit robust behavior without any additional engineering effort.

\begin{figure}[thpb]
  \centering
  \includegraphics[width=1\columnwidth]{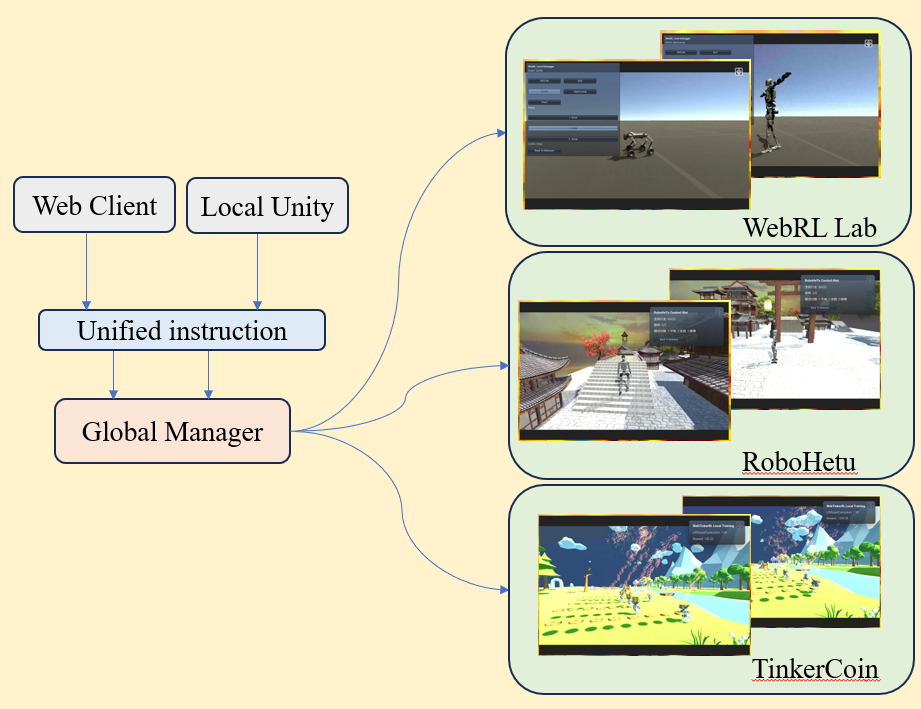}
  \caption{Unified scene routing in Web-Gewu. The \texttt{GlobalManager} decodes a unified command stream from both the Web client and the local Unity runtime, dynamically dispatching to one of three specialized educational scenes.}
  \label{fig:routing}
\end{figure}

The scene routing topology is illustrated in Fig.~\ref{fig:routing}. The \texttt{GlobalManager} acts as a central dispatcher: it accepts a unified instruction stream from both the browser client and the local Unity editor, decodes each command using the Envelope protocol, and dynamically routes it to the appropriate scene runtime---\emph{WebRL Lab} for policy validation and model switching, \emph{RoboHeTu} for complex terrain navigation, or \emph{TinkerCoin} for curriculum learning demonstration. This single-entry-point design ensures that scene transitions are atomic and idempotent regardless of the command source.

\section{RESULTS}

A video demo of Web-Gewu that demonstrates all functionalities is available at: \url{https://linqi-ye.github.io/video/web-gewu.mp4}

\subsection{Infrastructure Overhead}

A key design goal of Web-Gewu is to minimize cloud infrastructure cost by delegating all computation to the edge. To validate this, we measured the CPU utilization of the cloud signaling server (a 2-core vCPU, 2~GB RAM virtual machine running \texttt{frps} and \texttt{coturn}) during active sessions. Utilization remained stable at 0.8\%--1.1\% with a single active client, confirming that the cloud tier is a purely lightweight relay whose cost is independent of simulation complexity or fidelity. This contrasts sharply with conventional centralized cloud rendering, where GPU and bandwidth costs scale directly with the number of concurrent sessions and the rendering workload.

\subsection{Educational Sandbox Showcase}

\begin{figure}[thpb]
  \centering
  \includegraphics[width=\columnwidth]{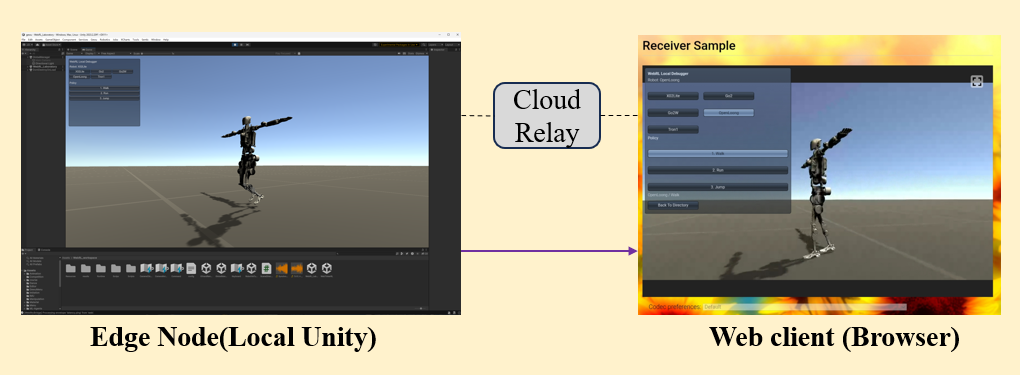}
  \caption{Cross-platform synchronization demonstration of the Web-Gewu educational sandbox. The browser-based Web client (right) abstracts the complex RL environment into an intuitive control panel and mirrors the high-fidelity simulation running on the local edge node (left) via the cloud relay with minimal end-to-end latency.}
  \label{fig:demo}
\end{figure}

Web-Gewu currently includes three built-in simulation scenarios covering representative robot morphologies and RL training paradigms. Fig.~\ref{fig:demo} illustrates the system in operation: the left panel shows the high-fidelity physics simulation running on the edge node inside the Unity editor, while the right panel shows the browser-based Web client receiving the mirrored live stream and presenting an intuitive control panel---all synchronized through the cloud relay with minimal perceptible latency.

\textbf{Playground.} A model inference showcase in which pre-trained RL policy weights for various robot platforms are loaded and their behaviors streamed in real time. Learners can interactively switch between trained models and observe policy differences in a controlled setting, providing an accessible entry point for understanding policy-driven behavior.

\textbf{RoboHeTu.} A navigation and locomotion demonstration featuring the G1 quadrupedal robot in a virtual environment inspired by the Song Dynasty painting \textit{Along the River During the Qingming Festival}. Users can switch between walk, run, and cross-terrain locomotion modes and issue omnidirectional movement commands via keyboard input, receiving real-time visual feedback of gait adaptation.

\textbf{TinkerCoin.} A live RL training demonstration in which the TinkerPro bipedal robot is trained from scratch using a physically grounded curriculum learning strategy. Inspired by the assistive force formulation proposed in HoST~\cite{hostpaper}, our curriculum applies a directional assistive force to the robot's torso, directed vertically upward with a 5° forward tilt relative to the robot's current heading--the tilt introduces a small tangential component that gently encourages forward locomotion without artificially constraining the policy.

The assistive force magnitude is governed by a scalar curriculum coefficient $\lambda \in [0, 1]$, where $\lambda = 1.0$ corresponds to an upward support force equal to 50\% of the robot's body weight ($F_{\uparrow} = \lambda \cdot 0.5\,mg$). The coefficient follows a piecewise-linear decay schedule:
\begin{equation}
  \lambda(t) = \max\!\left(0,\; 1.0 - 0.2\cdot\max\!\left(0,\left\lfloor\frac{t - 5\!\times\!10^5}{10^5}\right\rfloor\right)\right)
\end{equation}
That is, $\lambda$ remains at 1.0 for the first 500{,}000 training steps, then decreases by 0.2 for each subsequent 100{,}000 steps until it reaches 0.0 at step 1{,}000{,}000. Under this schedule, the robot first learns to balance and stride with substantial support, then progressively adapts to unsupported walking as the artificial aid is withdrawn.

\begin{figure}[thpb]
  \centering
  \includegraphics[width=\columnwidth]{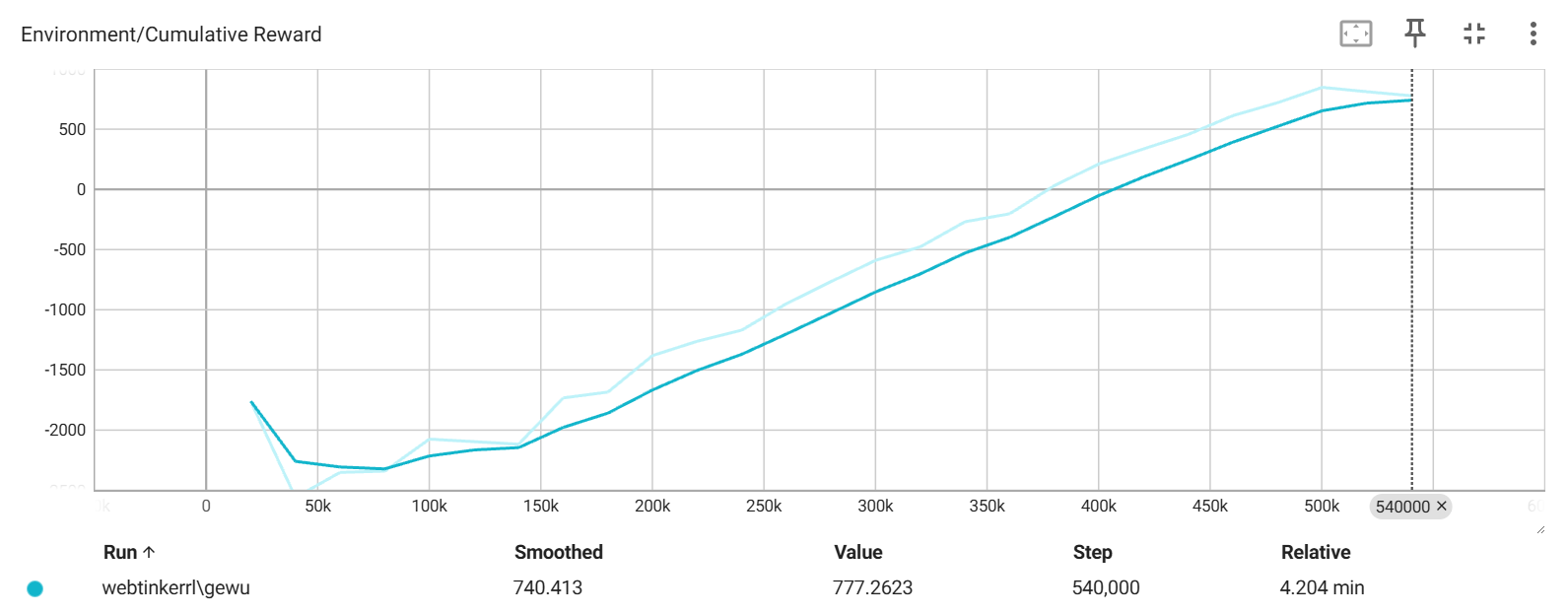}\\[3pt]
  \includegraphics[width=\columnwidth]{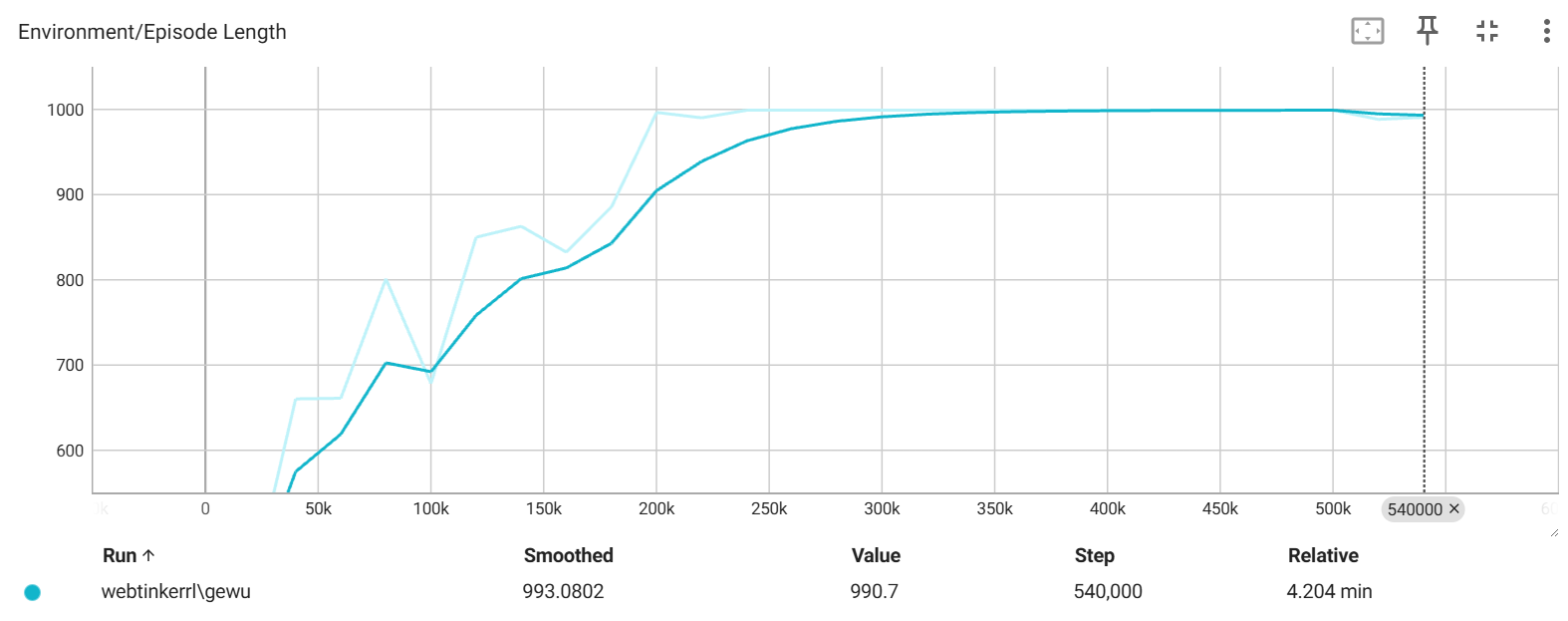}
  \caption{TinkerCoin training curves streamed live to the browser. \textit{Top}: cumulative reward over training steps. The reward dips initially as the agent explores with full episode length, then rises steadily as locomotion stabilizes. \textit{Bottom}: episode length converges to the maximum horizon (${\sim}1{,}000$) by ${\sim}250\text{k}$ steps, confirming sustained upright walking well before the curriculum support is fully withdrawn at step ${\sim}100\text{k}$.}
  \label{fig:training}
\end{figure}

In practice, the robot achieves stable forward locomotion in approximately 3~minutes of wall-clock time and fully unsupported continuous walking in approximately 8~minutes. The cumulative reward and episode length curves, streamed in real time to the browser and shown in Fig.~\ref{fig:training}, reveal the training dynamics clearly. The cumulative reward initially decreases as the agent begins using the full episode length and accumulates fall penalties, then rises monotonically once balance stabilizes, crosses zero near step 300{,}000, and reaches approximately $+600$ by step 550{,}000. The episode length curve rises from near zero to the maximum episode horizon (${\sim}1{,}000$ steps) by step 250{,}000, indicating that the robot sustains upright locomotion for full episodes well before the curriculum support is withdrawn. These real-time curves provide learners with an unambiguous, data-driven view of the exploration-to-exploitation transition central to RL.

Collectible coin assets are distributed throughout the scene; as the policy improves, the robot reaches and collects progressively more coins, providing a gamified progression that reinforces the concept of reward-driven policy improvement.

\subsection{End-to-End System Workflow}

The operational workflow is designed to minimize setup overhead for both instructors and learners. The cloud server (\texttt{frps} + \texttt{coturn}) runs as a persistent service requiring no per-session reconfiguration. On the edge side, the instructor launches \texttt{frpc} to expose the local signaling port on the public address, then starts the Unity runtime, which initializes to the \texttt{GlobalManager} bootstrap scene. Once these services are running, any learner can access the system by opening a single URL in a standard web browser. WebRTC ICE negotiation completes automatically, establishing either a direct P2P connection or a TURN-relayed fallback as network conditions dictate. The learner selects a scenario from the WebApp UI; \texttt{SceneDirector} handles scene transition, camera synchronization, and runtime binding without further manual intervention.

\section{CONCLUSIONS}

We have presented Web-Gewu, a browser-based interactive platform for robot reinforcement learning education built on a cloud-edge-client WebRTC collaborative architecture. By strictly delegating all physics simulation and RL computation to the edge node and confining the cloud tier to pure signaling relay, Web-Gewu resolves the long-standing tension between computational accessibility and high-fidelity simulation: a 2-core cloud server sustaining less than 1.1\% CPU utilization suffices to support multi-user sessions, while learners interact with complex robot simulations from any device capable of running a modern browser.

The Envelope JSON protocol and five-layer engineering architecture provide a robust, extensible foundation for bidirectional control and real-time telemetry visualization. The three built-in educational scenarios demonstrate the platform's ability to support both policy inference demonstrations and live from-scratch RL training, with real-time reward visualization serving as an intuitive bridge between abstract algorithm concepts and observable robot behavior. Beyond the three scenarios presented here, the framework's minimal scene integration contract---register, bind, and optionally add one typed handler---means that any new Unity-based robot scenario can be incorporated without modifying the transport, signaling, or protocol layers, positioning Web-Gewu as an open, community-extensible platform rather than a fixed demonstration tool.

Future work will focus on two directions: (1) opening a custom model weight upload interface to allow instructors and learners to deploy their own trained policies directly into the platform; and (2) integrating a hardware bridge to real-world robot, , enabling Sim2Real deployment demonstrations in which policies trained in Web-Gewu are transferred directly to physical hardware, closing the loop between simulation-based education and real-world embodied intelligence.

\addtolength{\textheight}{-12cm}   


\end{document}